\documentclass[acmtog]{acmart}
\usepackage{booktabs} % For formal tables

\usepackage[ruled]{algorithm2e} % For algorithms

\SetAlFnt{\small}
\SetAlCapFnt{\small}
\SetAlCapNameFnt{\small}
\SetAlCapHSkip{0pt}

\acmJournal{TOG}
\copyrightyear{2023} 
\acmYear{2023} 
\setcopyright{rightsretained} 
\acmConference[SIGGRAPH Conference Proceedings]{Special Interest Group on Computer Graphics and Interactive Techniques Conference Conference Proceedings}{Aug 6--10}{2023}
\acmBooktitle{Special Interest Group on Computer Graphics and Interactive Techniques Conference Conference Proceedings (SIGGRAPH '23 Conference Proceedings), August 6--10, 2023, Los Angeles, CA, USA}
\acmDOI{10.1145/3588432.3591494}
\acmISBN{979-8-4007-0159-7/23/08}

\usepackage{multirow}
\usepackage{subcaption}
\usepackage{graphicx}
\usepackage{mathtools}
\usepackage{tabularx}
\usepackage{pifont}

\graphicspath{{./figs/}}

\usepackage[normalem]{ulem}

\DeclareRobustCommand{\thinskip}{\hskip 0.1em\relax}
\def\emdash{---}
\def\d@sh#1#2{\unskip#1\thinskip#2\thinskip\ignorespaces}
\def\Dash{\d@sh\nobreak\emdash}
\def\Ldash{\d@sh\empty{\hbox{\emdash}\nobreak}}
\def\Rdash{\d@sh\nobreak\emdash}

\definecolor{good}{rgb}{1,1,0.6}
\definecolor{better}{rgb}{1,0.8,0.6}
\definecolor{best}{rgb}{1,0.6,0.6}

\newcommand{\cmark}{\textcolor{teal}{\ding{51}}}%
\newcommand{\xmark}{\textcolor{purple}{\ding{55}}}%

\begin{document}
\title{Single-Shot Implicit Morphable Faces with Consistent Texture Parameterization}

\author{Connor Z. Lin}
\affiliation{%
  \institution{Stanford University}
  \country{USA}
}
\affiliation{%
  \institution{NVIDIA}
  \country{USA}
}
\authornote{Work done during an internship at NVIDIA.}
\author{Koki Nagano}
\affiliation{%
  \institution{NVIDIA}
  \country{USA}
}
\author{Jan Kautz}
\affiliation{%
  \institution{NVIDIA}
  \country{USA}
}
\author{Eric R. Chan}
\authornotemark[1]
\affiliation{%
  \institution{Stanford University}
  \country{USA}
}
\affiliation{%
  \institution{NVIDIA}
  \country{USA}
}
\author{Umar Iqbal}
\affiliation{%
  \institution{NVIDIA}
  \country{USA}
}
\author{Leonidas Guibas}
\affiliation{%
  \institution{Stanford University}
  \country{USA}
}
\author{Gordon Wetzstein}
\affiliation{%
  \institution{Stanford University}
  \country{USA}
}
\author{Sameh Khamis}
\affiliation{%
  \institution{NVIDIA}
  \country{USA}
}

\begin{abstract}
There is a growing demand for the accessible creation of high-quality 3D avatars that are animatable and customizable. Although 3D morphable models provide intuitive control for editing and animation, and robustness for single-view face reconstruction, they cannot easily capture geometric and appearance details. Methods based on neural implicit representations, such as signed distance functions (SDF) or neural radiance fields, approach photo-realism, but are difficult to animate and do not generalize well to unseen data. To tackle this problem, we propose a novel method for constructing implicit 3D morphable face models that are both generalizable and intuitive for editing. Trained from a collection of high-quality 3D scans, our face model is parameterized by geometry, expression, and texture latent codes with a learned SDF and explicit UV texture parameterization. Once trained, we can reconstruct an avatar from a single in-the-wild image by leveraging the learned prior to project the image into the latent space of our model.
Our implicit morphable face models can be used to render an avatar from novel views, animate facial expressions by modifying expression codes, and edit textures by directly painting on the learned UV-texture maps. We demonstrate quantitatively and qualitatively that our method improves upon photo-realism, geometry, and expression accuracy compared to state-of-the-art methods.
\end{abstract}

\begin{CCSXML}
<ccs2012>
   <concept>
       <concept_id>10010147.10010371.10010382</concept_id>
       <concept_desc>Computing methodologies~Modeling/Geometry</concept_desc>
       <concept_significance>500</concept_significance>
       </concept>
 </ccs2012>
\end{CCSXML}

\ccsdesc[500]{Computing methodologies~Modeling/Geometry}

\keywords{Neural Avatars, Implicit Representations, Texture Maps, Animation, Inversion}

\begin{teaserfigure}
\centering
\includegraphics[width=\columnwidth]{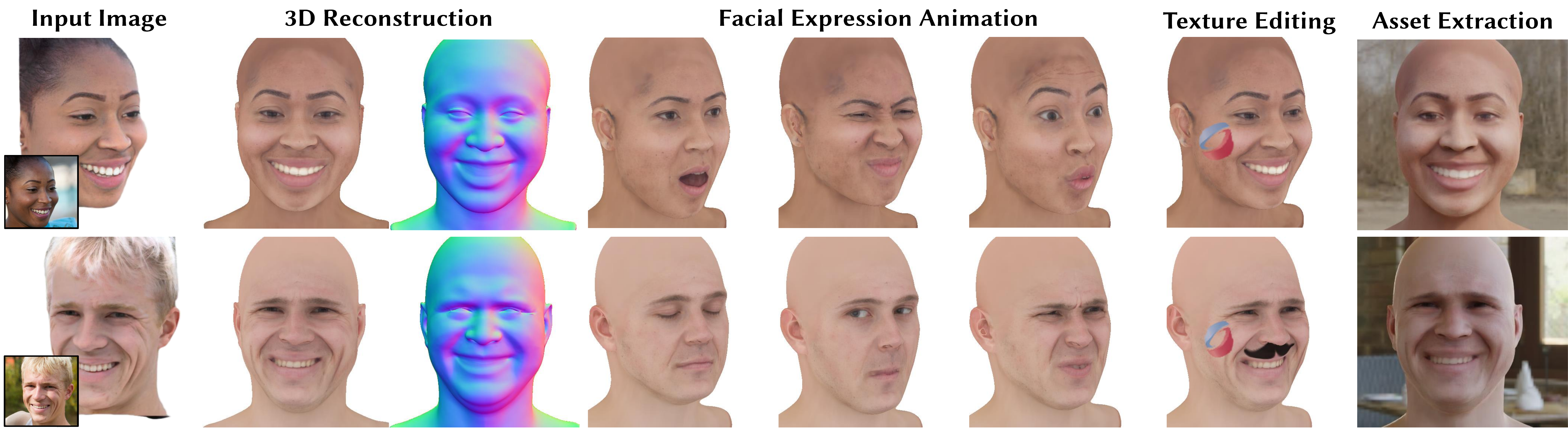}
\caption{Given a single input image, our method reconstructs a high-quality editable 3D digital avatar (columns 2 and 3) by combining implicit geometry representations with explicit texture maps. The proposed approach naturally supports novel view synthesis from large pose shifts, an expressive and non-linear facial animation space (columns 4 through 6), direct user access to texture map editing (column 7), and 3D asset extraction for further downstream applications such as relighting (column 8). Original image courtesy of COD Newsroom/flickr (top) and Malcolm Slaney/flickr (bottom).}
\label{fig:teaser}
\end{teaserfigure} 

\maketitle

\section{Introduction}
\label{sec:introduction}
Personalized avatar creation—the ability to map one's facial features to a 3D virtual replica that can be animated, customized, and rendered—is an emerging technology with great promise for cinema, the metaverse, and telepresence. Advances in this area may lead to digital twins with greater verisimilitude in detail and in animation that are more easily integrated into downstream applications and pipelines. Single-shot personalized avatar creation enables reconstructing face avatars from individual RGB images with greater convenience and flexibility than methods that require more specialized capture setups or procedures.

Traditional approaches to animatable 3D avatar creation are often based on 3D Morphable Models (3DMM)~\cite{blanz1999morphable}, which disentangle shape and appearance variation into a low-dimensional face representation. Building on these, more recent approaches often leverage either explicit (textured) template meshes~\cite{feng2021learning,danvevcek2022emoca,li2017learning,grassal2022neural,khakhulin2022realistic,tran2019learning} or neural implicit representations~\cite{park2019deepsdf,sitzmann2019scene,mildenhall2021nerf}. Template-based approaches enable easy asset extraction and intuitive editing, but are often unable to capture high-quality geometry and textures. Emerging implicit face models can achieve greater realism by modeling more complex geometric features such as hair~\cite{zheng2022avatar, cao2022authentic, giebenhain2022learning}. However, implicit face representations often compromise on interpretability and are less intuitive to control; the entangled latent spaces learned by these highly parameterized models are difficult to edit.

Our approach aims to combine the interpretability and editability advantages of template-based 3DMMs with the quality and topological flexibility of implicit 3D representations. Crucially, we decouple appearance and geometry into two branches of our network architecture. By incorporating a UV parameterization network to learn continuous and consistent texture maps, we can export avatars as textured meshes to support downstream applications such as texture map editing and relighting in a traditional graphics pipeline (See Figure~\ref{fig:teaser}). On the other hand, by representing geometry with an implicit signed distance field (SDF), our facial shape is less limited by resolution and topology compared to mesh-based approaches. 

We show that our proposed hybrid representation effectively captures the geometry, appearance, and expression space of faces. We demonstrate that single-shot in-the-wild portrait images can be effectively mapped to avatars based on our proposed representation, and that these avatars improve upon the previous state-of-the-art in photo-realism, geometry, and monocular expression transfer. Moreover, we demonstrate compelling capability for enabling direct texture editing and disentangled attribute editing such as facial geometry and appearance attributes.%extending the approach to handle the more complex topology of full human heads.

% In this work, we propose a novel hybrid framework representation that combines the high-quality geometry and flexible topology of implicit representations with direct control over the learned texture map.

In summary, contributions of our work include:
\begin{itemize}
    \item We propose a hybrid morphable face model combining the high-quality geometry and flexible topology of implicit representations with the editability of explicit UV texture maps. 
    \item We present a single-shot inversion framework to map a single in-the-wild RGB image to our implicit 3D morphable model representation. The inverted avatar supports novel view rendering, non-linear facial reanimation, disentangled shape and appearance control, direct texture map editing, and textured mesh extraction for downstream applications.
    \item We demonstrate state-of-the-art reconstruction accuracy for photo-realistic rendering, geometry, and expression accuracy in the single-view reconstruction setting.
    % \item We propose a novel hybrid representation that combines the high-quality geometry, non-linear animation space, and flexible topology of implicit representations with the interpretability of UV-parameterized texture maps. 
    % \item We present a single-shot inversion framework that can map an in-the-wild RGB image to our face representation. The inverted avatar supports novel view rendering, non-linear facial reanimation, and texture editing.
    % \item We demonstrate that our framework achieves state-of-the-art reconstruction accuracy for photo-realistic rendering, geometry, and expression accuracy.
    %\item We show that our learned consistent texture parameterization enables not only direct texture editing, but also a semantic mapping between 3D face geometry and 2D facial landmarks, which can be leveraged to improve single-shot reconstruction quality.
    %\koki{we can say that the consistent texture parameterization in implicit face models not only introduces the direct texture editing but also enables direct semantic mapping between the face model and 2D input (e.g., via facial landmarks), allowing significantly improved geometry reconstruction in the single-view reconstruction setting.}
\end{itemize}

\section{Related Work}
\label{sec:related}

\begin{table}[t]
    \caption{Comparison to recent prior work. To the best of our knowledge, our method is the first implicit 3D face model to generalize across single-image inputs while supporting flexible topology and explicit texture map control.}
    \resizebox{\columnwidth}{!}{
    \begin{tabular}{ l p{1.55cm} p{1.69cm} p{1.7cm} p{1.0cm}  }
    \toprule
        & Generalizable & Single-Image & Implicit\newline Representation & Explicit Texture Control \\\midrule
        EMOCA~\shortcite{danvevcek2022emoca}                                & \cmark & \cmark & \xmark & \cmark \\
        ROME~\shortcite{khakhulin2022realistic}                                 & \cmark & \cmark & \xmark & \xmark \\
        Neural Parametric Head Models~\shortcite{giebenhain2022learning}        & \xmark & \xmark & \cmark & \xmark \\
        IM-Avatar~\shortcite{zheng2022avatar}                            & \xmark & \xmark & \cmark & \xmark \\
        Neural Head Avatars~\shortcite{grassal2022neural}                  & \xmark & \xmark & \cmark & \cmark \\
        Volumetric Avatars from a Phone Scan~\shortcite{cao2022authentic} & \cmark & \xmark & \cmark & \cmark \\
        HeadNeRF~\shortcite{hong2022headnerf}                             & \cmark & \cmark & \cmark & \xmark \\
        \textbf{Ours}                        & \cmark & \cmark & \cmark & \cmark \\
    \bottomrule
\end{tabular}}
\end{table}

\subsection{Mesh-based 3D Morphable Models}
The seminal work by Blanz and Vetter proposed a linear 3D Morphable Model (3DMM)~\cite{blanz1999morphable} that models facial shape and textures on a template mesh using linear subspaces computed by principal component analysis (PCA) from 200 facial scans. This low-dimensional facial shape and texture space makes 3DMMs suitable for robustly capturing facial animation as well as reconstructing 3D faces in monocular settings. To reconstruct shape, texture, and lighting from a photo, previous work employed continuous optimization using constraints such as facial landmarks and pixel colors~\cite{Romdhani2005,Garrido2013,Shi2014,Cao2014,Ichim2015,thies2016face,Cao2016,Garrido2016,li2017learning} and more recently deep learning-based inference~\cite{Dou_2017_CVPR,Tran_2017_CVPR,Tewari_2017_ICCV,Genova_2018_CVPR,wu2019mvf,tewari2019FML,deng2019accurate,feng2021learning,danvevcek2022emoca,luo2021normalized,mbr_CFML,dib2021towards,dib2021practical}. While approaches relying on 3DMMs tend to be robust, they are ineffective for reconstructing high-fidelity geometry and texture details due to the linearity and low dimensionality of the model. Various other methods extended 3DMMs to capture non-linear shapes~\cite{chandran2020semantic,tran2018nonlinear,tran2019towards,tran2019learning,wang2022faceverse,li2020learning,tewari2018self}, photo-realistic appearance using neural rendering or optimization~\cite{thies2019neural,nagano2018pagan,Gecer_2019_CVPR,Saito_2017_CVPR}, or reflectance and geometry details for relightable avatar generation~\cite{lattas2020avatarme,Yamaguchi2018,Huynh_2018_CVPR,chen2019photo}. 
%normalized and relightable avatars\cite{}
%AvatarMe, DeepFacenormalization,Shichen paper,shugo paper
Recent approaches predict geometry offsets over the template mesh to reconstruct non-facial regions such as hair~\cite{grassal2022neural,khakhulin2022realistic}. We refer the reader to Egger et al.~\shortcite{egger20203d} for an in-depth survey of 3DMM techniques and Tewari et al.~\shortcite{tewari2022advances} for a report of recent advancements in neural rendering. 
% May add some discussion about the linear blendshapes for animation. 
% The first 3DMM~\cite{blanz1999morphable} modeled facial appearance and geometry using linear subspace deformations of a template mesh. Since then, recent works leverage convolutional encoders to map monocular images to their corresponding 3DMM parameters~\cite{li2017learning,feng2021learning,danvevcek2022emoca,deng2019accurate}. However, traditional 3DMMs are limited to linear expression animation and struggle with texture in occluded regions introduced during animation. Follow-up work has addressed many of these issues~\cite{chandran2020semantic,luo2021normalized, tran2018nonlinear,tran2019towards,tewari2018self,tran2019learning,bermano2014facial} and recent methods have utilized meshes produced by single-shot 3DMM encoders as an initialization step. The predicted mesh can be deformed to include regions such as hair~\cite{grassal2022neural, khakhulin2022realistic}, and higher quality dynamic textures can be synthesized using neural rendering~\cite{nagano2018pagan, doukas2021headgan}. We refer the reader to Egger et al.~\shortcite{egger20203d} for an in-depth survey of 3DMM techniques and Tewari et al.~\shortcite{tewari2020state} for a report of recent advancements in neural rendering. 

Since mesh-based 3DMMs represent geometry with a shared template mesh, their fixed topology limits the ability to scale the model to capture complex geometry such hair or fine-scale details. Additionally, their ability to synthesize photo-realistic facial textures may be limited by the resolution of the template mesh and discrete texture map. By parameterizing geometry with a signed distance function and color with a continuous texture map, our method is able to avoid such resolution issues and scale more efficiently with model capacity while retaining 3DMM-like intuitive parameters to individually control geometry and textures. Our consistent texture parameterization enables not only direct texture editing in UV space, but also semantic correspondence between our face model and an input image via facial landmarks, which can be leveraged to improve single-shot reconstruction quality.

%Mesh-based 3DMMs represent geometry with vertex deformations of an underlying template mesh. Therefore, their fixed topology limits the ability to capture complex geometry such as hair or accessories. Additionally, their ability to synthesize photo-realistic facial textures may be limited by the resolution of the template mesh and texture map. By parameterizing geometry with a signed distance function and color with a continuous texture map, our method is able to avoid such resolution issues and scale more efficiently with model capacity. Furthermore, we can place facial landmark constraints on the learned UV parameterization during training to explicitly provide the model with higher capacity in higher frequency regions such as the eyes or mouth interior.

\setlength{\belowcaptionskip}{-10pt}
\begin{figure*}
\begin{center}
    \includegraphics[width=2\columnwidth]{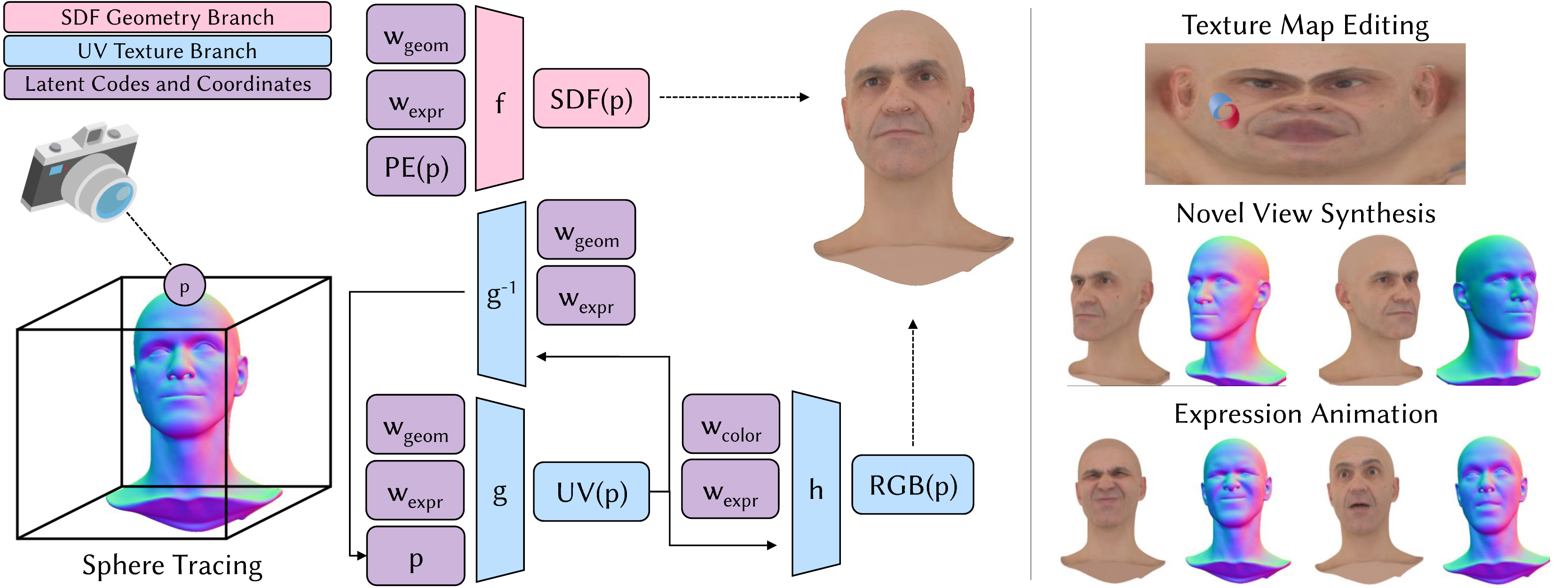}
\end{center}
   \caption{Our Pipeline. Avatars are represented by geometry, expression, and color latent codes $\{w_{geom}, w_{expr}, w_{color}\}$ with each being 512 dimensional. At each 3D coordinate $p$ during sphere tracing, the SDF network $f$ and UV parameterization network $g$ are conditioned on $w_{geom}$, $w_{expr}$, and positional encoding $PE(p)$ to predict the signed distance $SDF(p)$ and UV coordinates $UV(p)$, respectively. The inverse UV parameterization network $g^{-1}$ regularizes the learned mapping to be a surface parameterization $g^{-1}(UV(p); w_{geom}, w_{expr}) = p$, while the color network $h$ predicts the associated RGB texture $RGB(p) = h(UV(p); w_{color}, w_{expr})$. After training, the avatar can be rendered freely with direct control over its texture and facial expression, or extracted as a stand-alone textured mesh asset.}
\label{fig:pipeline}
\end{figure*}
\setlength{\belowcaptionskip}{0pt}

\vspace{-0.1in}
%\subsection{Implicit Representations for General 3D Objects}
\subsection{Implicit Representations for Modeling and Rendering}

While single-shot 3D reconstruction methods have explored various explicit 3D representations such as voxels~\cite{yan2016perspective, girdhar2016learning, zhu2017rethinking, tulsiani2017multi, wu2018learning, yang2018learning}, point clouds~\cite{fan2017point}, meshes~\cite{xu2019disn}, geometric primitives~\cite{zou20173d, niu2018im2struct}, and depth maps~\cite{wu2020unsupervised}, implicit representations have recently been leveraged to achieve higher resolution reconstruction using occupancy or signed distance fields (SDFs)~\cite{xu2019disn, mescheder2019occupancy, chen2019learning}. Implicit representations such as neural radiance fields (NeRFs)~\cite{mildenhall2021nerf} and signed distance fields (SDFs)~\cite{park2019deepsdf} have demonstrated high reconstruction quality for 3D shapes and volumetric scenes. PIFu~\cite{saito2019pifu} and follow-up works~\cite{cao2022jiff, saito2020pifuhd} use implicit fields to model human bodies and clothing. AtlasNet~\cite{groueix2018papier} demonstrated 3D shape generation by predicting a set of parametric surface elements given an input image or point cloud. NeuTex~\cite{xiang2021neutex} replaces the radiance prediction of NeRFs with a learned UV texture parameterization conditioned on lighting direction. Although our method also employs a UV cycle consistency loss, we 1) operate in a SDF setting and condition our parameterization on geometry and expression latent codes to generalize across samples rather than overfit to a single scene, 2) employ sparse facial landmark constraints to facilitate learning a semantically intuitive and consistent parameterization, and 3) explicitly leverage 2D to 3D facial landmark correspondences enabled by the learned consistent parameterization during single-image reconstruction. Implicit representations have also given rise to higher quality 3D generative models~\cite{chan2022efficient, or2022stylesdf, xue2022giraffe}, and follow-up work has studied inverting an image into the latent space of a pre-trained 3D GAN~\cite{roich2022pivotal, lin20223d, ko20233d} for single-view 3D reconstruction. However, without careful optimization and additional priors~\cite{xie2022high,yin2022spi}, this 3D GAN inversion tends to be less robust due to unknown camera poses~\cite{ko20233d} and multi-view nature of NeRF training in the monocular setting. %training the ambiguity of optimizing a volumetric representation on single-shot input. 
On the other hand, the compact face representation of our model provides robust initialization in the single-shot reconstruction setting.  %However, the limited pose distribution of 3D GAN image datasets~\cite{karras2019style} often causes the quality of rendered side views to degrade significantly. 

% Despite achieving higher reconstruction quality than their explicit counterparts, implicit representations are typically less intuitive to control beyond modifying a conditional latent code capturing texture or geometry. We propose a novel hybrid representation that combines SDFs with UV parameterizations to get the best of both worlds: high quality geometry with flexible topology while allowing for direct control over the learned texture map.

\subsection{Implicit Face Models}
Compared to traditional mesh-based 3DMMs for face modeling, implicit representations naturally offer flexible topology and non-linear expression animation through latent code conditioning. While some approaches learn to reconstruct an implicit 3DMM from an input 3D face scan \cite{zheng2022imface, yenamandra2021i3dmm, alldieck2021imghum, zanfir2022phomoh, cao2022authentic, giebenhain2022learning}, other works have explored modeling an implicit face model from RGB videos~\cite{zheng2022avatar, grassal2022neural, zheng2022pointavatar, ma2022neural}. However, the above approaches either do not support or cannot generalize to single-shot in-the-wild images. Multi-view methods have also been used to reconstruct implicit head models~\cite{kellnhofer2021neural,athar2022rignerf,hong2022headnerf, athar2021flame, li2022implicit,ramon2021h3d,wang2022morf}. HeadNeRF~\cite{hong2022headnerf} is the closest to our work and learns a parametric head model from multi-view images during training; at test-time, an input image can be inverted for 3D reconstruction. However, HeadNeRF performs volumetric rendering at a limited image resolution and relies on upsampling CNN modules, resulting in flickering artifacts from depth error during novel view synthesis. Furthermore, existing implicit morphable models do not support texture manipulation beyond interpolation; by contrast, our learned explicit texture paramterization enables intuitive and out-of-domain edits such as adding tattoos or mustaches (see Fig.~\ref{fig:teaser}). %
% Furthermore, existing implicit models are typically less intuitive for explicit texture manipulation beyond latent code interpolation; by contrast, our formulation, which incorporates an explicit UV parameterization, makes direct texture editing much more intuitive.

% \eric{Optionally include additional subsection that covers neural face representations that do *not* directly fall into mesh-3DMM/implicit-3DMM, e.g. DeepVoxels, EG3D, the one that breaks up faces into a bunch of blocks (followup to deepvoxels)}

\section{Method}
\label{sec:methods}
\subsection{Implicit Morphable Face Parameterization}

%\textit{Representation.}
We disentangle each facial avatar into identity and expression, where identity is encoded by geometry and color latent codes while expression is captured by an expression latent code. To attain both high-quality geometry and interpretable texture, our model consists of an implicit geometry branch and a UV texture parameterization branch. The geometry branch contains a multilayer perceptron (MLP) that maps 3D points $p$ to SDF values $SDF(p)$ during sphere tracing. The UV texture branch consists of a parameterization MLP that maps $p$ to spherical coordinates $UV(p)$, a parameterization regularizer MLP that learns the inverse mapping from $UV(p)$ back to $p$, and a color network that predicts the output RGB at $UV(p)$. See Figure~\ref{fig:pipeline} for a diagram of our model pipeline. Please refer to the supplement for model architecture details.

%\connor{justification for not using renderpeople and for tripleganger. we need volume and quality for the purpose of single view 3D reconstruction. } 
%\textit{Dataset and Training.}
We train our model on the Triplegangers~\shortcite{tripleganger} 3D scan dataset for its volume and diversity of subjects and expressions. Although the RenderPeople~\shortcite{renderpeople} dataset additionally models hair and clothing, it only contains 120 neutral expression subjects, making it less suitable for reconstructing an avatar from unconstrained in-the-wild photos. Our training samples consist of a 3D head mesh, UV diffuse texture map, and six diffusely lit frontal RGB images. The dataset contains 515 different subjects each with 20 expressions, for a total of 10,300 data samples. Our full model learns an AutoDecoder dictionary of 515 geometry codes, 515 color codes, and 10,300 expression codes, as subjects express the same sentiment differently. Different expressions for the same training subject share the same geometry and color codes, allowing the model to disentangle expression from the underlying geometry and texture. Please refer to the supplement for examples of our training data.

%During training, we optionally constrain all subjects to share the same expression code for their neutral expression; this allows us to invert an in-the-wild subject and more easily animate them to a neutral expression.

\subsection{Training Losses}
Our model is trained on geometry, color, and regularization losses:
\begin{align}
    \mathcal{L} &= \mathcal{L}_{geom} + \mathcal{L}_{color} + \mathcal{L}_{reg}
\end{align}

Following Figure~\ref{fig:pipeline}, let $f$ be the SDF MLP, $g$ the UV parameterization MLP, $g^{-1}$ the inverse UV parameterization MLP, and $X$ the set of randomly sampled surface points during training. The geometry loss consists of surface, Eikonal~\cite{gropp2020implicit}, normal, and UV losses. The surface loss $\ell_{surf}$ optimizes the SDF zero level set, the Eikonal loss $\ell_{eikonal}$ regularizes the SDF gradients, and the normal loss $\ell_{normal}$ aligns the SDF gradients with the ground truth mesh normals $\hat{n}$. The UV loss $\ell_{uv}$ regularizes the learned mapping to follow an invertible surface parameterization, which enables correspondences between texture and geometry used in our single-shot inversion pipeline, described in Section 3.5.
\begin{align}
    \ell_{surf} &= \frac{1}{|X|}\sum_{x \in X}|f(x)| \\
    \ell_{eikonal} &= \mathbb{E}_{x}(\|\nabla_xf(x)\| - 1)^2 \\
    \ell_{normal} &= \frac{1}{|X|}\sum_{x \in X}\| \nabla_{x}f(x) - \hat{n}(x) \|^2 \\
    \ell_{uv} &= \frac{1}{|X|}\sum_{x \in X}\| x - g^{-1}(g(x)) \|^2 \\
    \mathcal{L}_{geom} &= \ell_{surf} + \ell_{eikonal} + \ell_{normal} + \ell_{uv}
\end{align}

The color loss consists of a reconstruction loss $\ell_{tex}$ on the ground truth texture $\hat{T}$, as well as perceptual~\cite{zhang2018unreasonable} and reconstruction losses $\ell_{img}$ over the facial region $I_{face}$ between the ground truth image $\hat{I}$ and rendered image $I$ obtained via sphere tracing:
\begin{align}
    \ell_{tex} &= \frac{1}{|X|}\sum_{x \in X}\| \hat{T}(x) - h(g(x)) \|^2 \\
    \ell_{img} &= LPIPS(\hat{I}_{face}, I_{face}) + \|\hat{I}_{face}- I_{face}\|^2 \\
    \mathcal{L}_{color} &= \ell_{tex} + \ell_{img} 
\end{align}

Finally, we enforce the compactness in the learned latent space by penalizing the magnitude of the geometry, color, and expression codes:
\begin{equation}
    \mathcal{L}_{reg} = \| w_{geom} \|^2 + \| w_{color} \|^2 + \| w_{expr} \|^2
\end{equation}

\begin{figure}
\begin{center}
    \includegraphics[width=0.95\columnwidth]{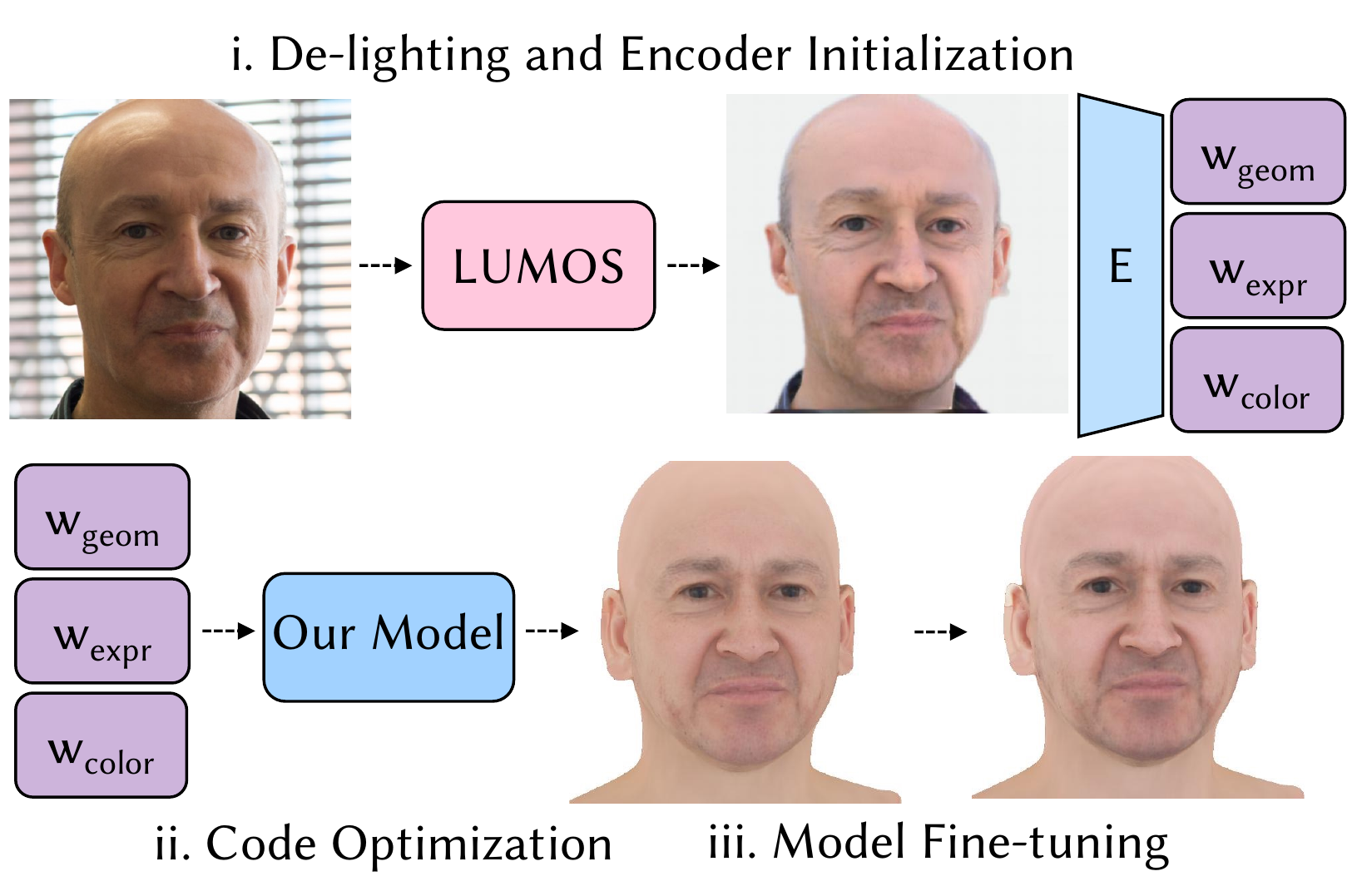}
\end{center}
   \caption{Single-shot inversion pipeline. We de-light the input image and initialize the latent codes using a pre-trained encoder (top row). We then perform PTI~\cite{roich2022pivotal} to get the final reconstruction (bottom row). Original image courtesy of Brett Jordan/flickr.}
\label{fig:inversion}
\end{figure}

\subsection{Learning UV Parameterizations}
\begin{figure}[t]
\begin{center}
    \includegraphics[width=\columnwidth]{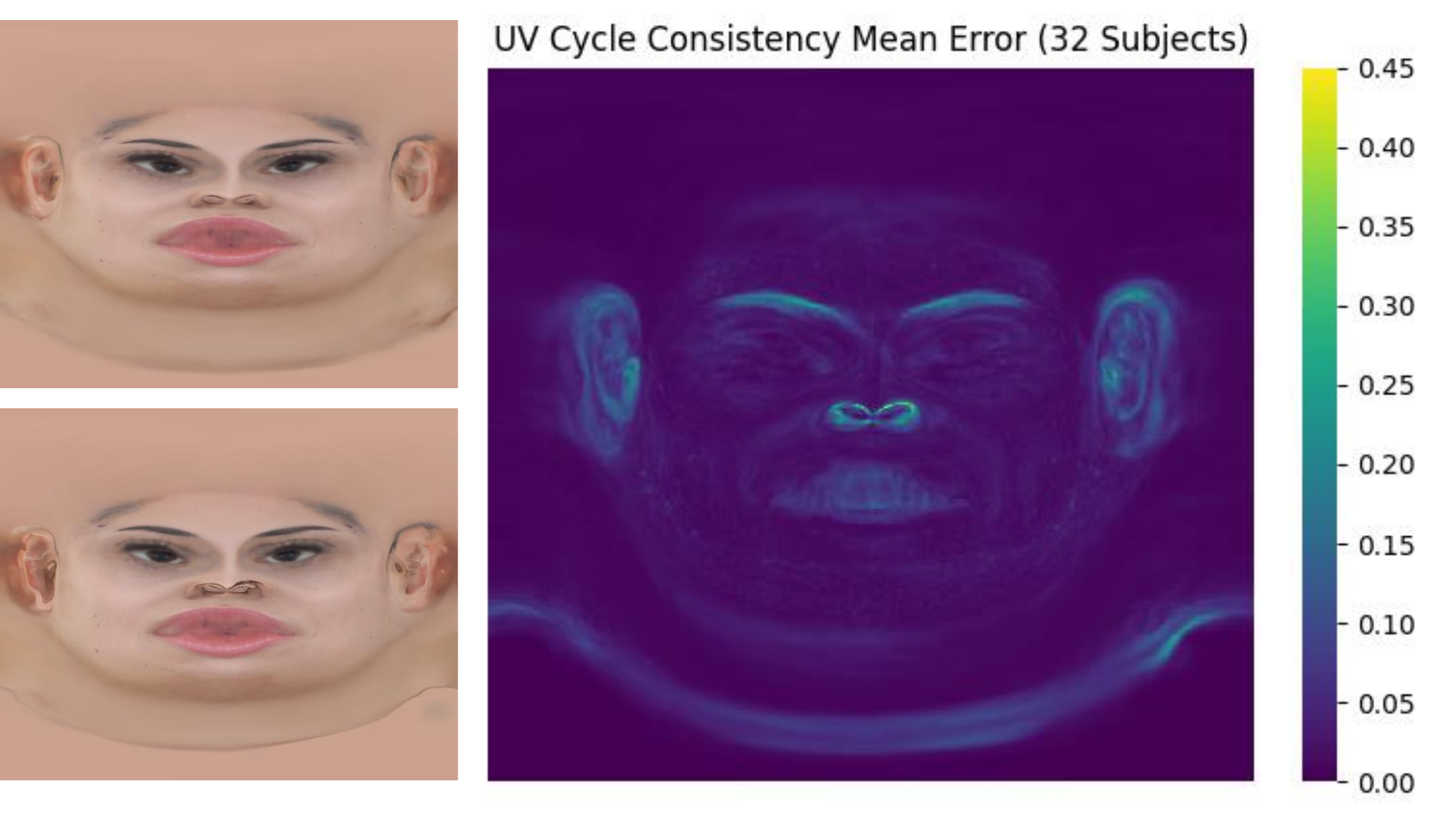}
\end{center}
    \vspace{-6pt}
   \caption{UV parameterization consistency. We measure the mean L2 error over 32 FFHQ subjects between the learned texture map (top left) and the cycle texture map (bottom left) obtained by mapping from UV $\rightarrow$ 3D $\rightarrow$ UV.}  
   \label{fig:uv_consistency}
\end{figure}

%\koki{If we keep this as a subsection it has to be something like "Learning UV Parameterization". If it is about loss functions, we can keep it under Training losses and downgrade this to paragraph.}
To learn an interpretable texture space and coherent semantic correspondence across subjects, we add an auxiliary loss term to $\mathcal{L}_{reg}$ that enforces the parameterization to be consistent through a sparse set of facial landmark constraints:
%In order for the learned UV mapping to enable texture editing and correspondences between 2D texture and 3D facial geometry, we regularize the parameterization to be consistent by enforcing a sparse set of facial landmark constraints during training:
\begin{align}
    \ell_{landmark} = \frac{1}{|L|}\sum_{x \in L}\| \hat{g}(x) - g(x) \|^2 + \| x - g^{-1}(g(x)) \|^2 
    \label{eq:landmarks}
\end{align}
The first term enforces the learned UV mapping to match the ground truth UV mapping $\hat{g}$ for the set of 3D facial landmark points $L$, and the second term enforces this mapping to be invertible. Fig.~\ref{fig:uv_consistency} demonstrates the consistency of our learned UV parameterization. Although mostly consistent, it is difficult to obtain perfect registrations around the inner mouth and eyes due to the billboard geometry and errors originating from the ground truth data.

%The learned semantic correspondences allow our method to align the predicted facial landmarks with those of the input image to improve single-shot reconstruction. See Section 3.5 for more details.

%We compute $L$ for each training image by extracting 5 landmarks from MTCNN and 60 landmarks from dlib. These 2D landmarks are then projected onto the surface of the corresponding 3D mesh, where $\hat{g}$ can then be computed via barycentric interpolation. Note that $\hat{g}$ can be scaled to directly control the network capacity for representing facial features, such as to capture higher frequency details in the eyes and mouth regions.

\subsection{Animation}
\begin{figure}
\begin{center}
    \includegraphics[width=\columnwidth]{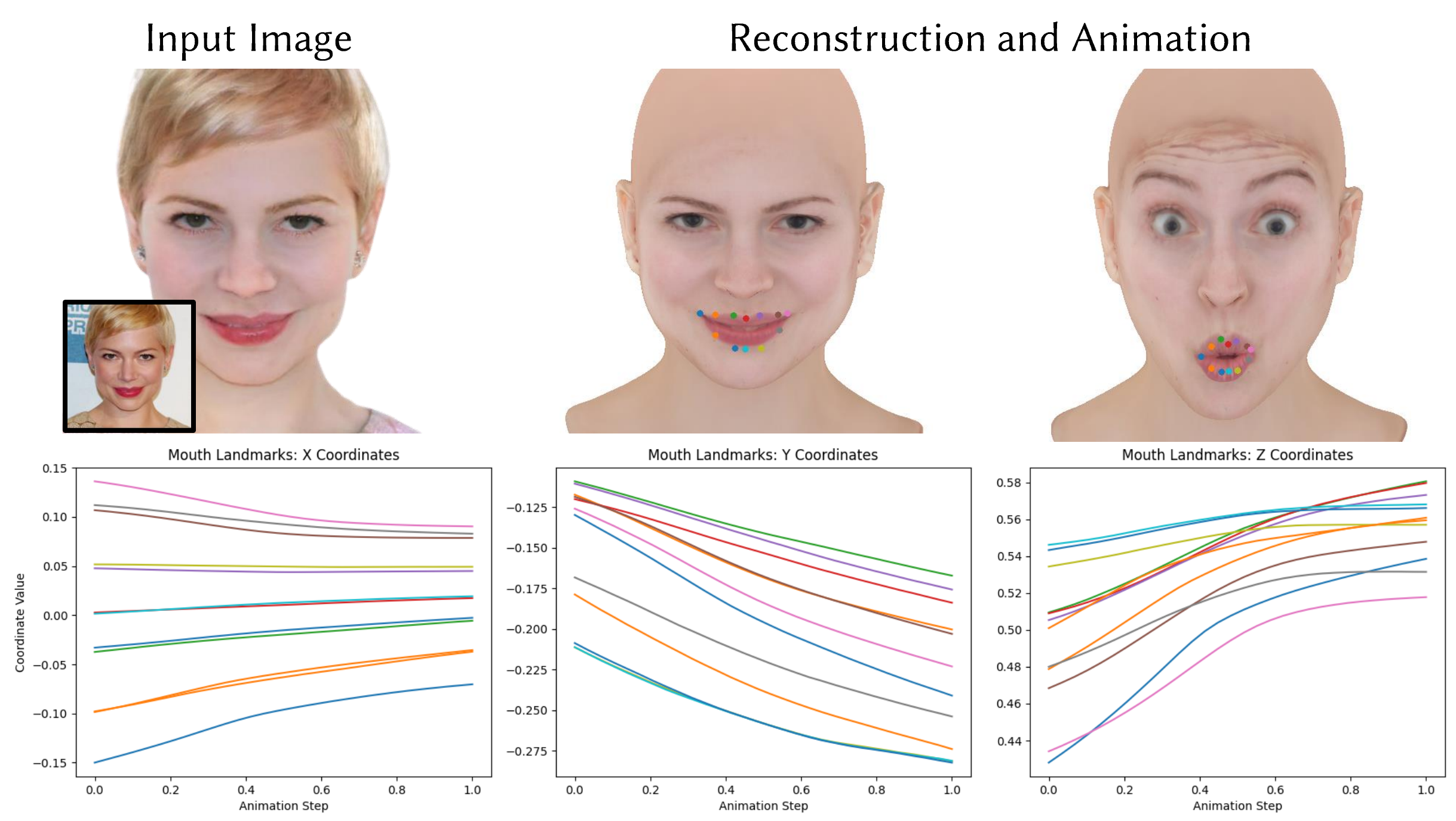}
\end{center}
   \caption{Non-linear animation space. By linearly interpolating between source and target expression codes, our model exhibits non-linear deformation trajectories on the 3D mouth vertices visualized. Original image courtesy of 
David Shankbone/flickr.
}  
   \label{fig:nonlinear}
\end{figure}

%\koki{forward cite the non linear animation. update the equation.}
After training, an avatar can be animated by manipulating its expression latent code. For a source subject with expression code $w_{expr}$, target expression code $w_{expr}'$, and animation timesteps $t \in [0, 1]$, we define the expression animation trajectory by:
\begin{equation}
w_{expr}(t) = w_{expr} + t * (w_{expr}' - w_{expr})
\end{equation}
Unlike traditional linear 3DMM approaches, our expression space follows non-linear trajectories learned from high-quality 3D scans, as shown in Fig.~\ref{fig:nonlinear}.

\subsection{Single-Shot Inversion}
\begin{figure}[t]
\centering
% \begin{center}
    \includegraphics[width=\columnwidth]{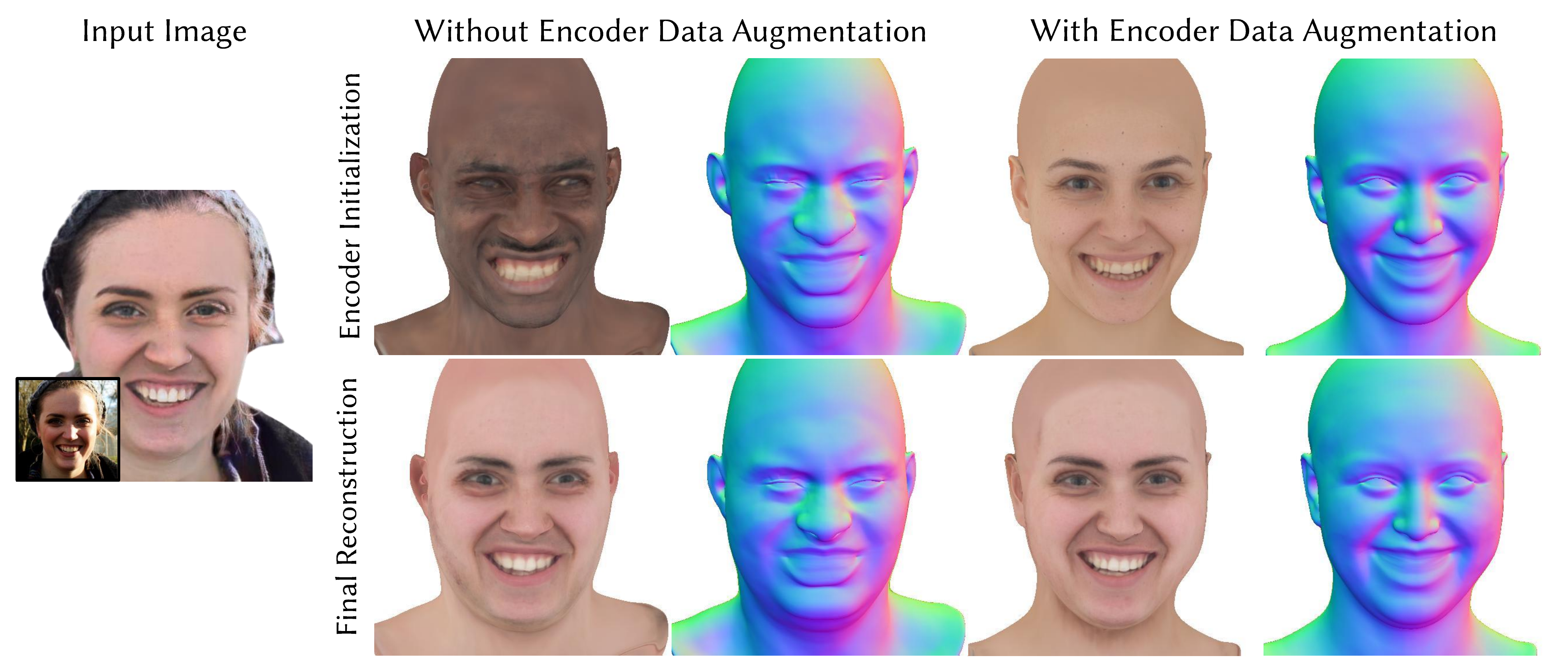}
% \end{center}
   \caption{Encoder training data augmentation ablation. Training the encoder with the synthetically augmented Triplegangers dataset ~\cite{yeh2022learning} significantly improves our initialization, which is important for converging to a high quality inversion result. Note the difference in the final reconstructed geometry. Original image courtesy of David Geitgey Sierralupe/flickr.}
   \label{fig:lumos_ablation}
\end{figure}

In order to reconstruct and animate unseen subjects, we project an input RGB image into the latent space of our pre-trained model and lightly fine-tune the model weights similar to Pivotal Tuning Inversion (PTI)~\cite{roich2022pivotal}. To handle unseen lighting conditions, we de-light the input image using LUMOS~\cite{yeh2022learning} and initialize the geometry, color, and expression codes through a separately trained encoder. We empirically find this encoder initialization to be important in obtaining robust results for in-the-wild input images (See Figure~\ref{fig:lumos_ablation}). 

\paragraph{Image Encoder.} We attain latent code initializations by training a DeepLabV3+~\cite{chen2018encoder} encoder to reconstruct each training image $\hat{I}$ and its corresponding latent codes $\hat{W}$ already computed from the previous AutoDecoder training stage:
\begin{align}
    \mathcal{L}_{enc} &=  \| \hat{I} - I \|^2 + \|\hat{W} - W \|^2 \\
    W &= [w_{geom}; w_{color}; w_{expr}] 
\end{align}
 One major challenge when inverting in-the-wild images is handling unseen identities, accessories, hairstyles, and occlusion present in real-world images, as Triplegangers contain limited identities with no variations in hairstyles or background. Therefore, we augment the encoder's training dataset with synthetically augmented Triplegangers images from ~\cite{yeh2022learning}, which improves the robustness of the initialization and final inversion reconstruction, shown in Fig.~\ref{fig:lumos_ablation}. 

\begin{figure}
\begin{center}
    \includegraphics[width=\columnwidth]{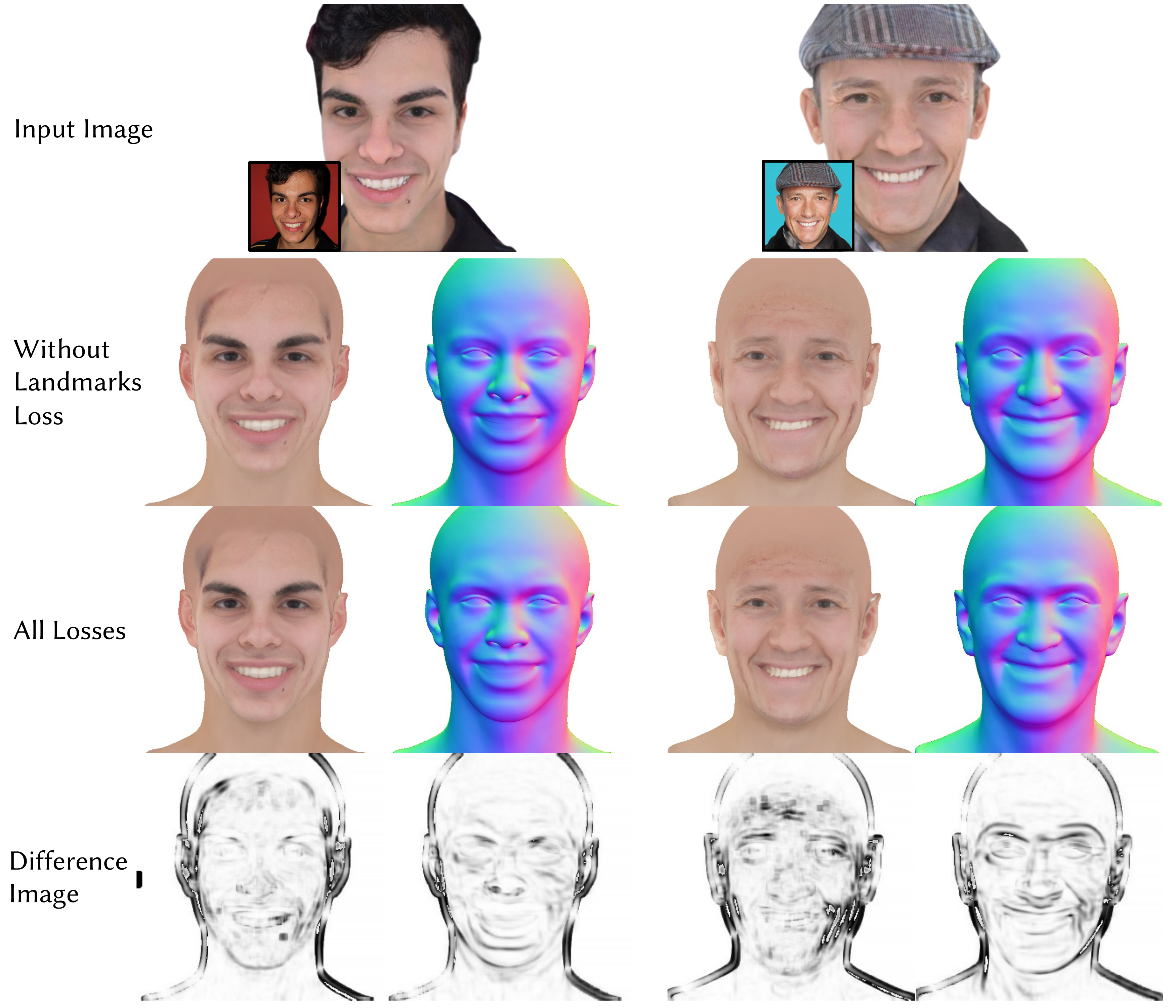}
\end{center}
   \caption{Facial landmarks loss ablation. Removing the facial landmarks loss during inversion reduces reconstruction quality of the face contour (left and right jaws) and facial features such as the eyes (right). Original image courtesy of Cena Mineira (left) and BigBrother Junkie (right).}
   \label{fig:losses_ablation}
\end{figure}

\paragraph{Optimization.} After initializing the latent codes for an input image $\hat{I}$ using our encoder, we freeze the model weights and optimize the latent codes while minimizing image, silhouette, multi-view consistency, facial landmark, and regularization losses:
\begin{align}
    \ell_{img} &= LPIPS(\hat{I}_{face}, I_{face}) + \| \hat{I}_{face} - I_{face} \|^2 \\
    \ell_{silhouette} &= \!\!\! \sum_{x \in \hat{I}_{face} \wedge x \not\in I_{face}} \!\!\! f(x) \\
    \ell_{ID} &= ArcFace(\hat{I}, I, I_{rand}) \\
    \ell_{landmark} &= \sum_{d \in D(\hat{I})}\| d - proj_{2D}(g^{-1}(\hat{d}))\|^2 \\
    \ell_{reg} &= \| w_{geom} \|^2 + \| w_{color} \|^2 + \| w_{expr} \|^2
\end{align}
where the silhouette loss $\ell_{silhouette}$ iterates over points contained in the ground truth face region $\hat{I}_{face}$, but not in the predicted face region $I_{face}$, to bring the points closer to the SDF zero level set. ArcFace~\cite{deng2019arcface} measures the face similarity between different views and $I_{rand}$ is a predicted render from a randomly perturbed camera pose. $D$ is an off-the-shelf facial landmark detector~\cite{dlib09} and $\hat{d}$ is the ground truth facial landmark UV mapping enforced in Eq.~\ref{eq:landmarks}. Note that our consistent UV parameterization directly enables correspondences for the facial landmark alignment loss $\ell_{landmark}$; Fig.~\ref{fig:losses_ablation} demonstrates the benefits of incorporating this loss. The regularization loss $\ell_{reg}$ is important to ensure that the optimized codes stay near the manifold of the pre-trained latent space for expression animation. We obtain face masks using a pre-trained BiSeNet~\cite{yu2018bisenet} and optimize for 800 steps.
%\koki{I think landmark loss should not be taken for granted because it is possible only because we have this consistent parameterization.
%if the results without landmarks are significantly worse, then it means that our UV learning offers more than just texture editing, but important for single-view reconstruction.}

\paragraph{Fine-tuning.} To reconstruct finer details in the input image, we freeze the latent codes after optimization and fine-tune the model weights on the above losses. We omit the silhouette loss, as we find it tends to bloat the geometry when the model weights are unfrozen. Although fine-tuning the model improves reconstruction quality, it may also hinder its capability for animation or novel view synthesis. Therefore, we only perform model fine-tuning for 60 steps.

\section{Results}
\label{sec:experiments}
\begin{figure*}[t]
    \centering
    \includegraphics[width=\textwidth]{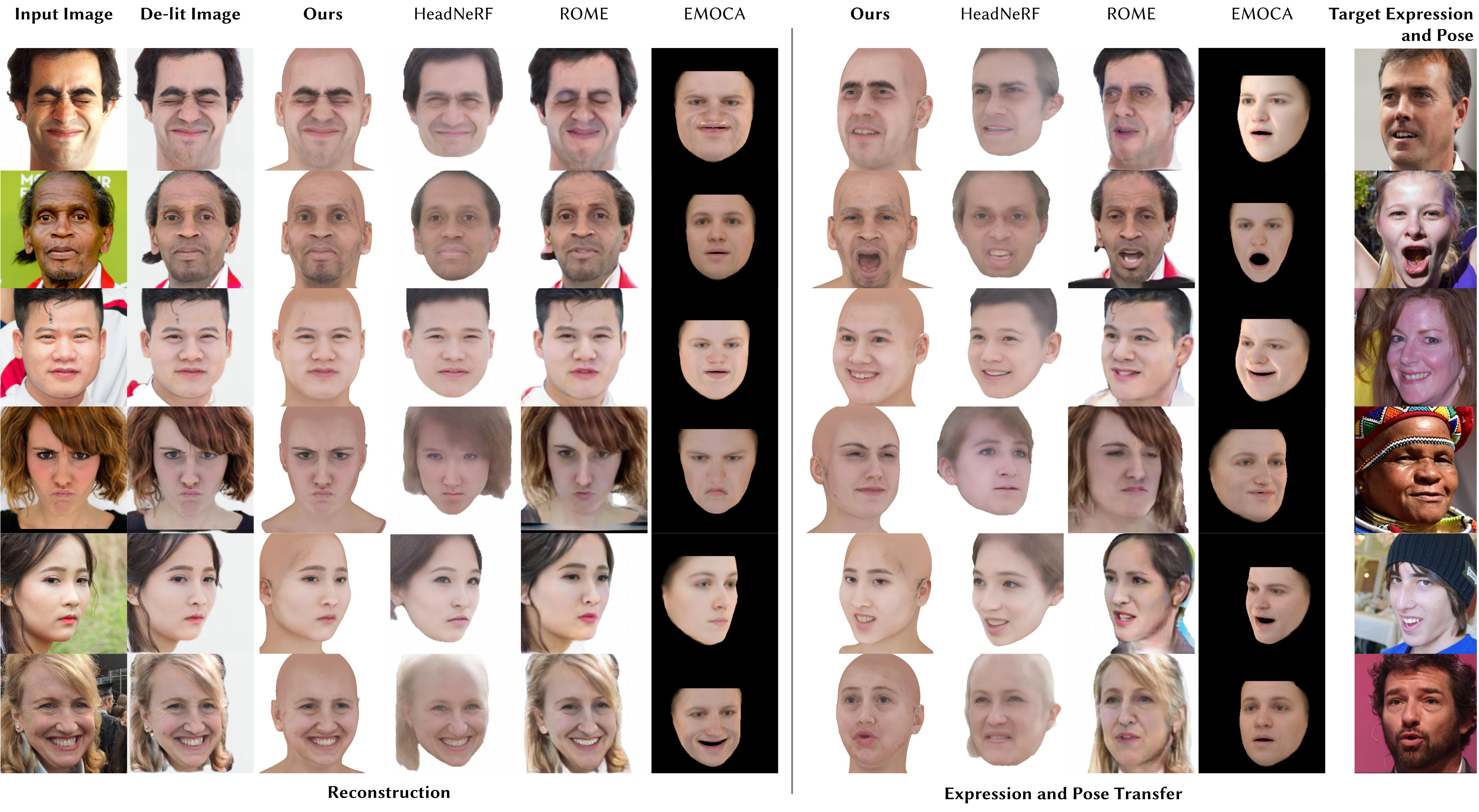}
    \caption{Single-shot reconstruction on FFHQ with expression and pose transfer. On the left, we show the input FFHQ source image, de-lit input image using LUMOS~\cite{yeh2022learning}, and reconstruction results for each method. On the right, we show monocular performance capture and retargeting, where we reconstruct and transfer the expression and pose from a target image (right-most column) to the source image identity (left-most column). On the left from top to bottom, original images are courtesy of José Carlos Cortizo Pérez/filckr, Montclair Film/flickr, Pham Toan/flickr, Javier Morales/flickr, Khiet Nguyen/flickr, and Malcolm Slaney/flickr. On the right from top to bottom, original images are courtesy of Adam Charnock/flickr, Daughterville Festival/flickr, Delaney Turner/flickr, South African Tourism/flickr, Pat (Cletch) Williams/flickr, and Collision Conf/flickr.}
    \label{fig:qualitative}
\end{figure*}

\begin{table*}[t]
    \centering
    \small
    \caption{Quantitative results on single-shot in-the-wild reconstruction (left) and self-expression retargeting (right). 
    \textbf{Left}: image, pose, and identity metrics are computed on 500 images sampled from FFHQ. Depth metrics are computed on the H3DS dataset. Image, identity, and depth metrics are computed only on the facial region. EMOCA is evaluated using its smaller face crop. 
    \textbf{Right}: FACS coefficients and facial landmarks are computed after expression and pose transfer on 32 expression pairs sampled from the Triplegangers test split.}
    \begin{tabular}{lccccc|p{1.0cm}p{1.0cm}}
    \toprule
        Reconstruction & LPIPS$\downarrow$  & DISTS$\downarrow$ & SSIM$\uparrow$ & Pose$\downarrow$ & ID$\uparrow$ & L1 Depth$\downarrow$ & RMSE Depth$\downarrow$ \\\midrule
        EMOCA & 0.1122 & 0.1268 & 0.9182 & 0.0681 & 0.0697 & 0.0300 & 0.0677 \\ 
        ROME & 0.1054  & 0.1130 & 0.9317 & 0.0600 & 0.3866 & 0.0237 & 0.0513 \\
        HeadNeRF  & 0.1090 & 0.1199 & 0.9268 & 0.0606 & 0.2334 & 0.0379 & 0.0695 \\ 
        Ours (optimization-free)  & 0.1427 & 0.1465 & 0.9053 & 0.0549 & 0.1082 & 0.0357 & 0.0658 \\
        Ours (encoder-free)  & 0.0890  & 0.0921 & 0.9441 & \textbf{0.0533} & 0.4600 & 0.0241 & 0.0527 \\
        Ours       & \textbf{0.0879}  & \textbf{0.0905}  & \textbf{0.9451} & 0.0563 & \textbf{0.4670} & \textbf{0.0228} & \textbf{0.0510} \\
    \bottomrule
    \end{tabular}
    \quad
    \begin{tabular}{lp{1.0cm}p{1.0cm}p{1.0cm}}
    \toprule
        Retargeting & FACS$\downarrow$  & Facial Landmarks$\downarrow$ \\\midrule
        EMOCA     & 4.712 & 0.2088 \\ 
        ROME      & 3.204 & 0.1414 \\
        HeadNeRF  & 3.848 & 0.1641 \\
        Ours      & \textbf{1.733} & \textbf{0.1165} \\
    \bottomrule
    \end{tabular}    
    \label{tab:metrics}
\end{table*}

\begin{figure}
\begin{center}
    \includegraphics[width=\columnwidth]{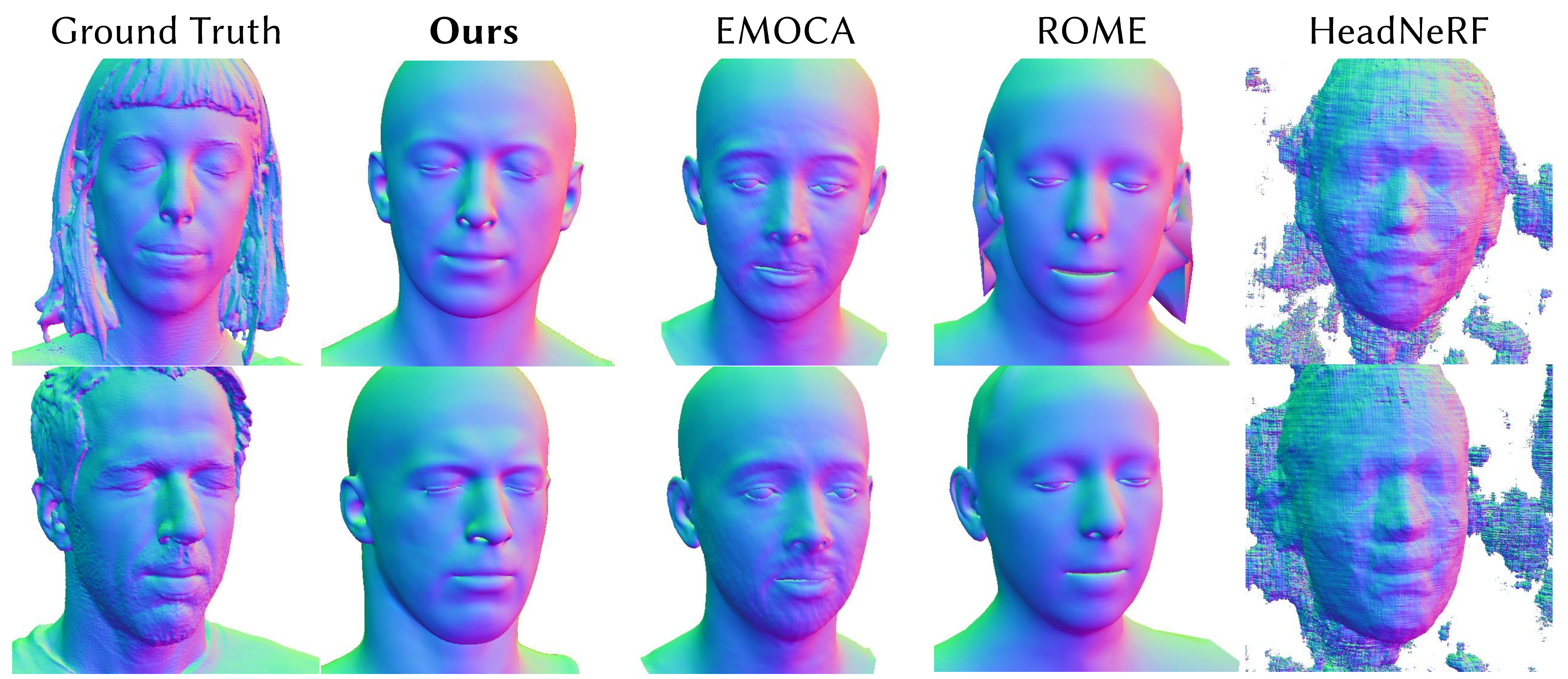}
\end{center}
   \caption{Ground truth geometry comparison on the H3DS dataset in the single-view setting.}
   \label{fig:shape}
   %Our method produces significantly better geometry than single-view implicit methods (HeadNeRF) and is competitive with mesh-based methods (EMOCA and ROME).
\end{figure}

\begin{table}[t]
    \caption{Quantitative comparison with FaceVerse~\cite{wang2022faceverse} on 500 sampled FFHQ images for single-shot in-the-wild reconstruction.}
    \resizebox{0.7\columnwidth}{!}{
    \begin{tabular}{ l c c c}
    \toprule
        Reconstruction & LPIPS$\downarrow$  & DISTS$\downarrow$ & SSIM$\uparrow$ \\\midrule
        FaceVerse         & 0.1280 & 0.1119 & 0.9126 \\
        Ours              & \textbf{0.0879} & \textbf{0.0905} & \textbf{0.9451} \\
    \bottomrule
\end{tabular}}
\label{tab:metrics2}
\end{table}

We present results of our proposed method with comparisons to EMOCA~\cite{danvevcek2022emoca}, ROME~\cite{khakhulin2022realistic} and FaceVerse~\cite{wang2022faceverse}, three recent mesh-based approaches for single-shot 3D avatar generation, and HeadNeRF~\cite{hong2022headnerf}, an implicit approach using neural radiance fields.
% EMOCA adds an expression shape encoder to DECA~\cite{feng2021learning} and encodes the input image into identity, expression, pose, and albedo parameters to reconstruct a coarse 3D mesh. The coarse mesh is then refined by a decoder conditioned on the regressed expression parameters. ROME encodes the input image into a neural texture and uses DECA to predict an initial mesh, which is further refined by an encoder conditioned on the neural texture. The final rendering is achieved by rendering the neural texture using a convolutional U-Net. HeadNeRF conditions a neural radiance field on identity, expression, albedo, and illumination parameters extracted from NL3DMM~\cite{tran2018nonlinear} to reconstruct the input image.
Our method achieves higher fidelity texture and geometry reconstruction in the facial region compared to the baselines. Qualitatively and quantitatively, our method also demonstrates more faithful expression and pose transfer between in-the-wild source and target images. Finally, our learned texture map is intuitive to edit and propagates naturally during animation.

\subsection{Implementation Details}
Our model is trained in two stages. In the first stage, we withhold the ground truth multi-view images, as we find that supervising with both texture maps and multi-view images negatively impacts the model's ability to learn a consistent UV mapping. In the second stage, we freeze the UV networks $\{g, g^{-1}\}$ and supervise using the multi-view images to fine-tune the learned texture maps while rendering image reconstructions at 768 $\times$ 512 resolution. Camera poses are provided with ground truth training data and we estimate camera poses for in-the-wild FFHQ images using Deep3DFaceRecon~\cite{deng2019accurate}. We perform sphere tracing for 50 steps per ray and use a dimensionality of 512 for the geometry, color, and expression latent codes. We train our AutoDecoder for 1000 epochs (approx. one week) and our inversion encoder for 200 epochs (approx. one day) across 8 NVIDIA A40 GPUs. We use a Triplegangers training/test split of 386/129 for the quantitative expression experiments. Sphere tracing takes 8.5 seconds and inversion takes 3 hours per image. See supplemental material for more details on training and model architectures. 

\subsection{Single-Shot 3D Face Reconstruction and Animation}
%We evaluate our proposed method for single-shot 3D face reconstruction and animation by comparing against EMOCA~\cite{danvevcek2022emoca}, ROME~\cite{khakhulin2022realistic}, HeadNeRF~\cite{hong2022headnerf}, and two ablations for our method.

\begin{figure}
    \includegraphics[width=\columnwidth]{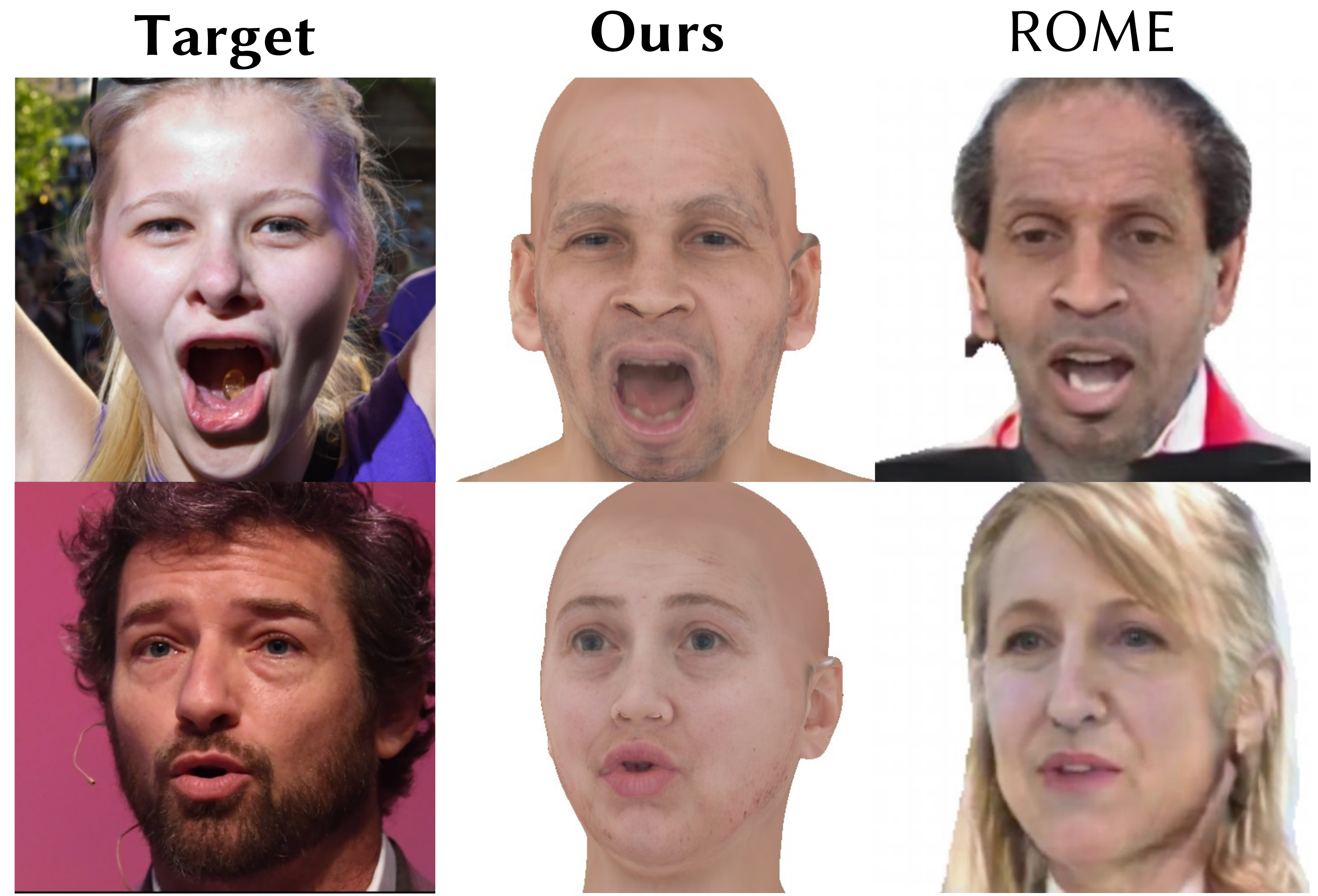}
    \caption{Zoomed in comparison with ROME~\cite{khakhulin2022realistic} from Fig.~\ref{fig:qualitative}. Our model captures the target expression with higher fidelity and higher resolution textures (512$\times$512) compared to ROME (256$\times$256).}
    \label{fig:qualitative2}
\end{figure}

\paragraph{Qualitative Results}
We show qualitative comparisons for single-shot reconstruction followed by expression and pose transfer on FFHQ~\cite{karras2019style} images between the proposed method, EMOCA, ROME, and HeadNeRF in Fig.~\ref{fig:qualitative} and Fig.~\ref{fig:qualitative2}.

Overall, our method is more photo-realistic and achieves higher expression accuracy in facial reconstruction. EMOCA does not model the mouth interior and relies on a pre-trained FLAME~\cite{li2017learning} albedo model for texture. 
%For expression and pose transfer, we selected target images that differ significantly from the input reconstructed image in expression, pose, gender, and/or ethnicity to test the disentanglement of each method. 
Our model produces the most faithful expression transfer, demonstrating the diversity of its learned expression space and generalization capabilities of our method to in-the-wild data. HeadNeRF exhibits a large amount of identity shift during pose transfer, whereas our method remains view-consistent after large pose changes. 

We also show a ground truth comparison of reconstructed geometry on the H3DS~\cite{ramon2021h3d} dataset between our method and the baselines in Fig.~\ref{fig:shape}. HeadNeRF performs volumetric rendering at a low resolution and therefore produces noisy depth results. Our geometry captures higher fidelity facial geometry than ROME and captures the expression more faithfully (e.g., eye blink) compared to EMOCA. 
%Our facial geometry is on par with or better than EMOCA and ROME without requiring a 3D template mesh prior for each input.

\paragraph{Quantitative Results}
We report quantitative reconstruction and self-reenactment expression transfer results in Table~\ref{tab:metrics} and Table~\ref{tab:metrics2}.
The photometric (LPIPS~\cite{zhang2018unreasonable}, DISTS~\cite{ding2020image}, SSIM~\cite{wang2004image}), pose error, and MagFace~\cite{meng2021magface} identity consistency (ID) metrics are calculated over a dataset of 500 images from FFHQ. We compute L1 and RMSE depth error over all subjects in the H3DS dataset. To evaluate self-reenactment expression error, we randomly sample 32 source--target expression pairs over a test split of the Triplegangers dataset and measure the L2 error for FACS~\cite{ekman1978facial} coefficients and facial landmarks. For details related to how each metric is computed, please refer to the supplemental material.

% should we mention here that although technically TG only has 20 expression categories, people express the same emotion in very different ways?

On the FFHQ dataset, our proposed method achieves the best accuracy in terms of LPIPS, DISTS, SSIM, and ID score. The optimization-free ablation struggles to handle the considerably large domain shift between Triplegangers training data and FFHQ in-the-wild images. Our model also exhibits the lowest depth error on the H3DS dataset without relying on a 3D template mesh prior. Finally, our model has the lowest FACS and facial landmark errors, demonstrating the diversity of its learned expression space.

\subsection{Ablations}
In addition to the baselines mentioned, we compare our method to two ablations for single-shot reconstruction. The first ablation is an optimization-free inversion approach that only uses the learned encoder to directly map an input image to the geometry, color, and expression codes $\{w_{geom}, w_{color}, w_{expr}\}$. The second ablation is an encoder-free inversion approach that omits the encoder and instead uses a mean initialization for $\{w_{geom}, w_{color}, w_{expr}\}$ over the learned AutoDecoder dictionary of latent codes.

Quantitative results for the ablations are reported in Table~\ref{tab:metrics}. The optimization-free approach produces significantly worse photometric and depth results, as there is a large domain gap between Triplegangers training data and in-the-wild images; this causes the encoder to produce a coarse reconstruction. The encoder-free approach performs better than the optimization-free approach but is still worse than our full method in image and geometry quality, demonstrating that the encoder initialization improves the optimization reconstruction. Both ablations and our full method perform similarly on pose accuracy.

\begin{figure}
\begin{center}
    \includegraphics[width=\columnwidth]{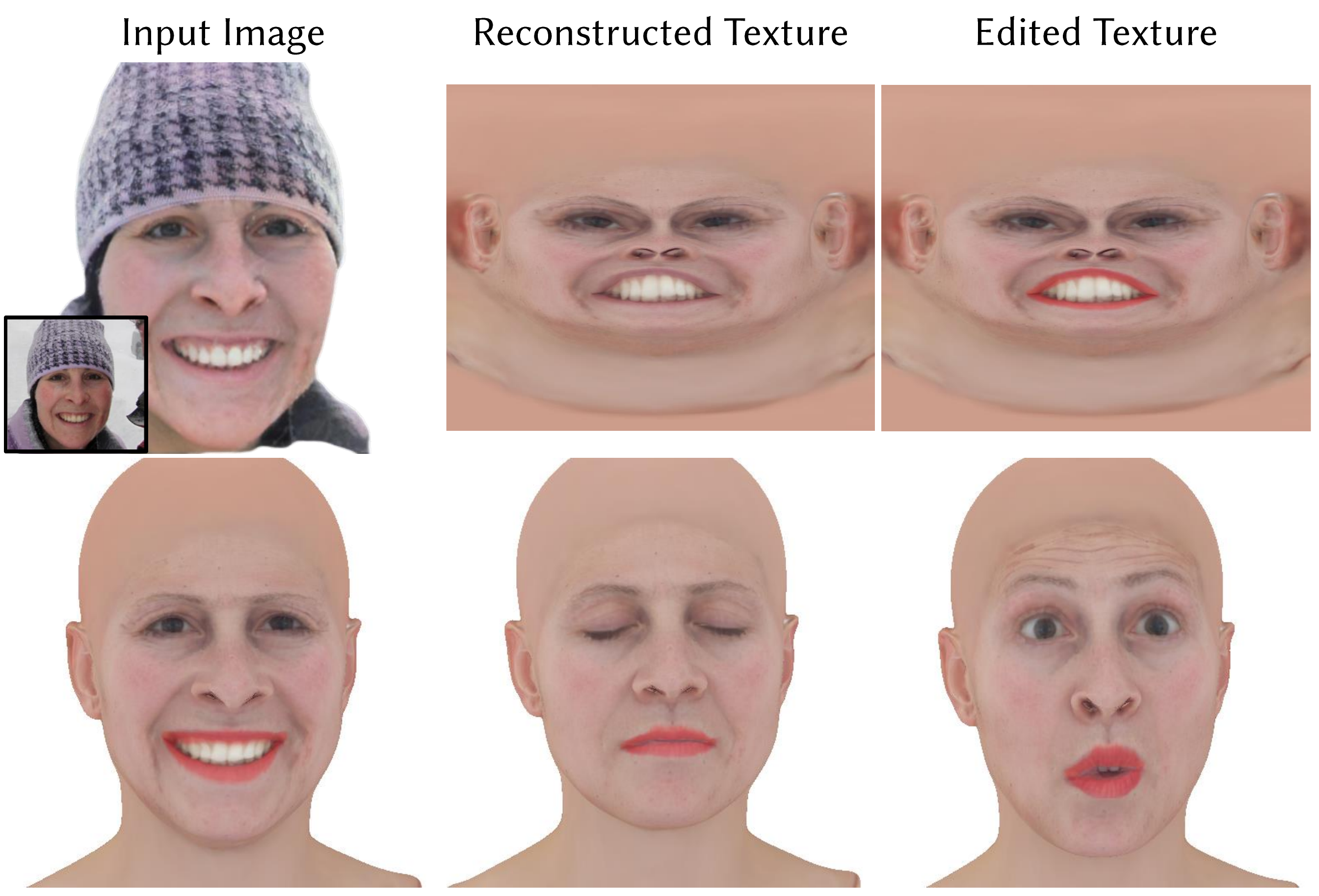}
\end{center}
   \caption{Texture editing. Top row: input image, learned texture map, and user edited texture map. The learned texture map layout is intuitive and edits propagate naturally during facial animation as shown in the bottom row. Original image courtesy of Ed Kohler/flickr.}
   \label{fig:tex_editing}
\end{figure}

\begin{figure}[b]
\begin{center}
    \includegraphics[width=\columnwidth]{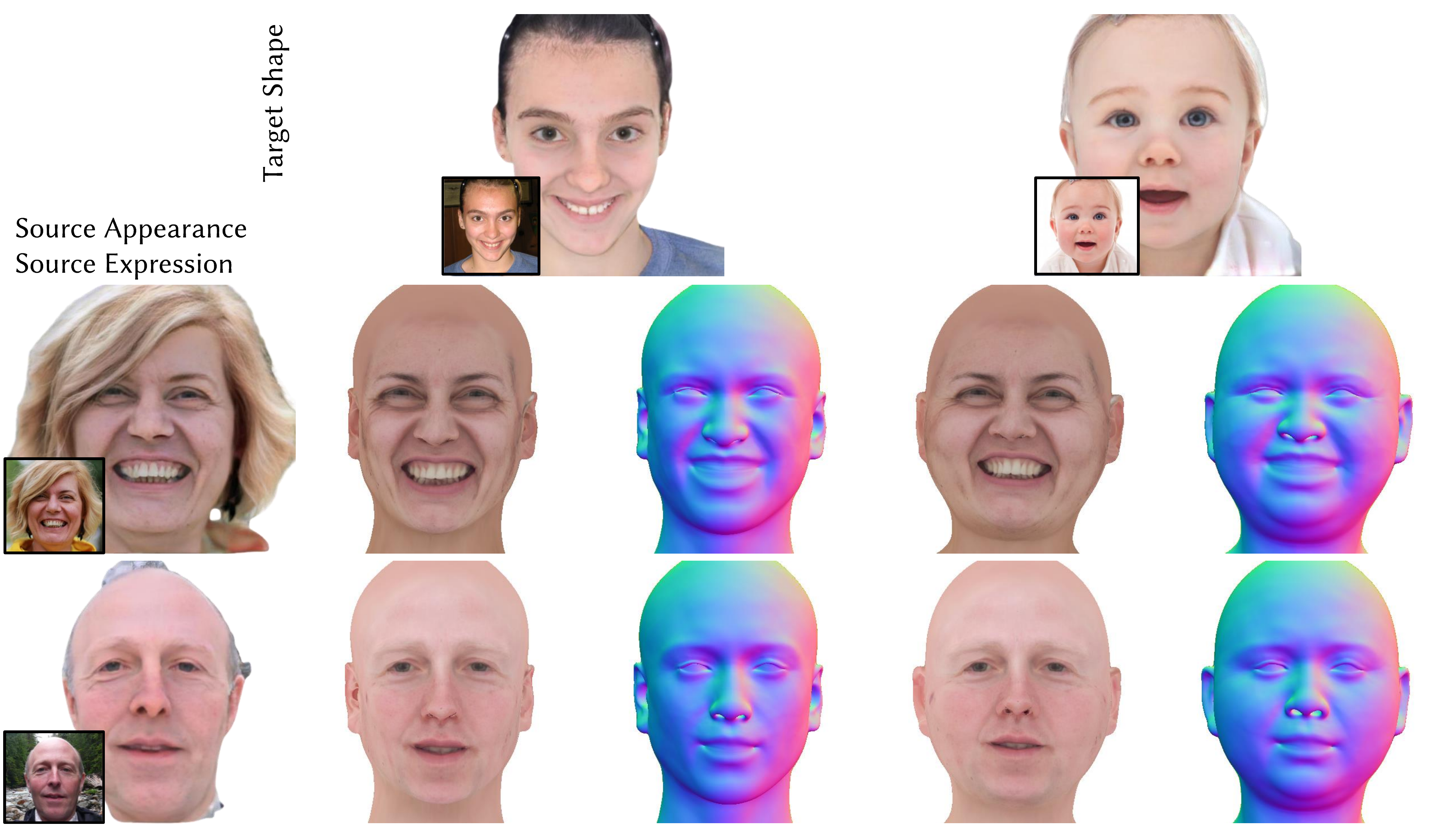}
\end{center}
   \caption{Shape attribute transfer. We fix the color and expression codes for the source subject and directly replace the source geometry code with the target geometry code. Original images are courtesy of Francesco Pierantoni/flickr (left col, top), Tim Regan (left col, bottom), Bob n Renee/flickr (top row, left), and Sarah \& Austin Houghton-Bird/flickr (top row, right).}
   \label{fig:shape_transfer}
\end{figure}

\begin{figure}[b]
\begin{center}
    \includegraphics[width=\columnwidth]{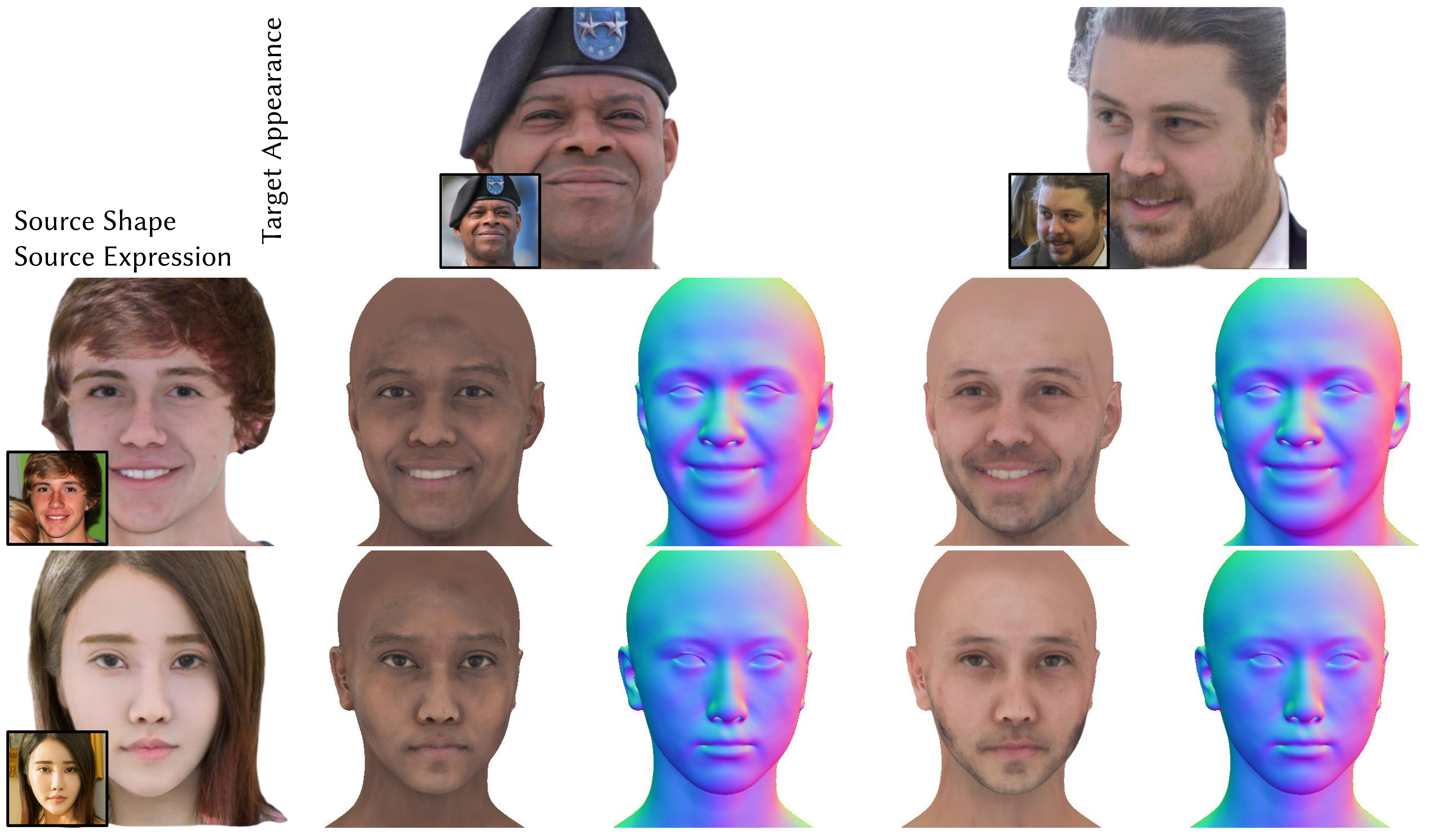}
\end{center}
   \caption{Facial appearance attribute transfer. We fix the geometry and expression codes for the source subject and directly replace the source color code with the target color code. Original images are courtesy of Lord Jim/flickr (left col, top), xiǎo cháo zhù/flickr (left col, bottom), U.S. Army/flickr (top row, left), and U.S. Department of Energy/flickr (top row, right).}
   \label{fig:color_transfer}
\end{figure}

\paragraph{Applications}
As demonstrated in Fig.~\ref{fig:qualitative}, our method directly supports monocular facial performance capture and expression retargeting. Our hybrid representation provides direct control over an intuitive texture map with a consistent layout. Fig.~\ref{fig:tex_editing} demonstrates an example workflow: a user reconstructs an input image and modifies the learned texture map. The edits then continue to persist smoothly across different facial animations. Textured meshes can be extracted for further downstream applications such as re-lighting, as shown in the teaser. Fig.~\ref{fig:shape_transfer} and Fig.~\ref{fig:color_transfer} further demonstrate our model's disentanglement between geometry, texture, and expression with its capability of shape and facial appearance transfer.

%\vspace{-0.1in}
% \begin{figure}
% \begin{center}
%     \includegraphics[width=\columnwidth]{SIGGRAPH2023/figs/nonlinear.pdf}
% \end{center}
%    \caption{Non-linear animation space. We demonstrate the non-linearity of our learned animation space by linearly interpolating between source and target expression codes while tracking the 3D trajectory of facial landmarks.}  
%    \label{fig:nonlinear}
% \end{figure}

\section{Discussion}
\label{sec:discussion}
We have presented a new method for reconstructing 3D animatable and textured faces from a single RGB image. The proposed approach combines implicit representations with explicit texture maps to support explicit editing while achieving better photo-realistic rendering, geometry, and expression reconstruction than previous methods. We believe the proposed method makes important contributions towards accessible creation of high-fidelity avatars from in-the-wild images that are animatable, editable, and customizable for downstream applications. 
%The proposed method makes important contributions towards improving the quality and control of geometry, textures, and expressions for 3D faces reconstructed from single-shot images. 

However, there are still limitations to the method. Firstly, the current optimization process during inversion is significantly slower than encoder-based methods. For real-time applications, more expressive representations such as neural feature fields can be explored to enable optimization-free inversion methods. Furthermore, the method relies on a de-lighting module from Lumos to process in-the-wild images to generate a diffusely lit input image, which  may cause subjects to appear paler than expected. These limitations may be alleviated through lighting augmentations of the training dataset to reduce the domain gap and incorporating a lighting model such as spherical harmonics into the representation. Finally, the results shown in this paper do not capture hair or accessories due to limitations of the training dataset. 
%Unfortunately, there does not currently exist such a substantially large and diverse 3D human scan dataset, but 
While not perfect, we refer to the supplemental material for a preliminary demonstration of our representation's capacity to handle hair and clothing on the smaller RenderPeople dataset. As implicit representations such as neural radiance fields excel at capturing the geometry and texture of thin structures, it may be fruitful to combine our method with recent sparse view implicit hair models~\cite{wu2022neuralhdhair, kuang2022deepmvshair}. 

%The proposed method makes important contributions towards improving the quality and control of textures and expressions for 3D faces reconstructed from single-shot images. 

%We believe this work to be particularly relevant in synthesizing virtual avatars suitable for social media, telepresence, and cinema.

\begin{acks}
 We thank Simon Yuen and Miguel Guerrero for helping with preparing the 3D scan dataset and assets, and Ting-Chun Wang for providing Lumos code. We also thank Nicholas Sharp, Sanja Fidler and David Luebke for helpful discussions and supports. This project was supported in part by a David Cheriton Stanford Graduate Fellowship, ARL grant W911NF-21-2-0104, a Vannevar Bush Faculty Fellowship, a gift from the Adobe corporation, Samsung, and Stanford HAI.
\end{acks}

\clearpage

% Bibliography
\bibliographystyle{ACM-Reference-Format}
\bibliography{references}

\clearpage 
\label{sec:figures_only}
\textcolor{white}{.}

%\koki{Try stylized inputs. } \koki{show avatar normalization (neutralized expression and lighting) in the application.}

\begin{comment}
\begin{figure}[b]
\begin{center}
    \includegraphics[width=\columnwidth]{SIGGRAPH2023/figs/extraction.pdf}
\end{center}
   \caption{The extracted assets (head mesh and texture) can be rendered in DCC tools like Blender for re-lighting and avatar customization with hair card models. Please see teaser for the extracted head meshes without hair.}
   \label{fig:extraction}
\end{figure}
\end{comment}

\setlength{\abovecaptionskip}{0pt}
\setlength{\belowcaptionskip}{-5pt}

\setlength{\belowcaptionskip}{-10pt}
\begin{figure*}
\begin{center}
    \includegraphics[height=575pt]{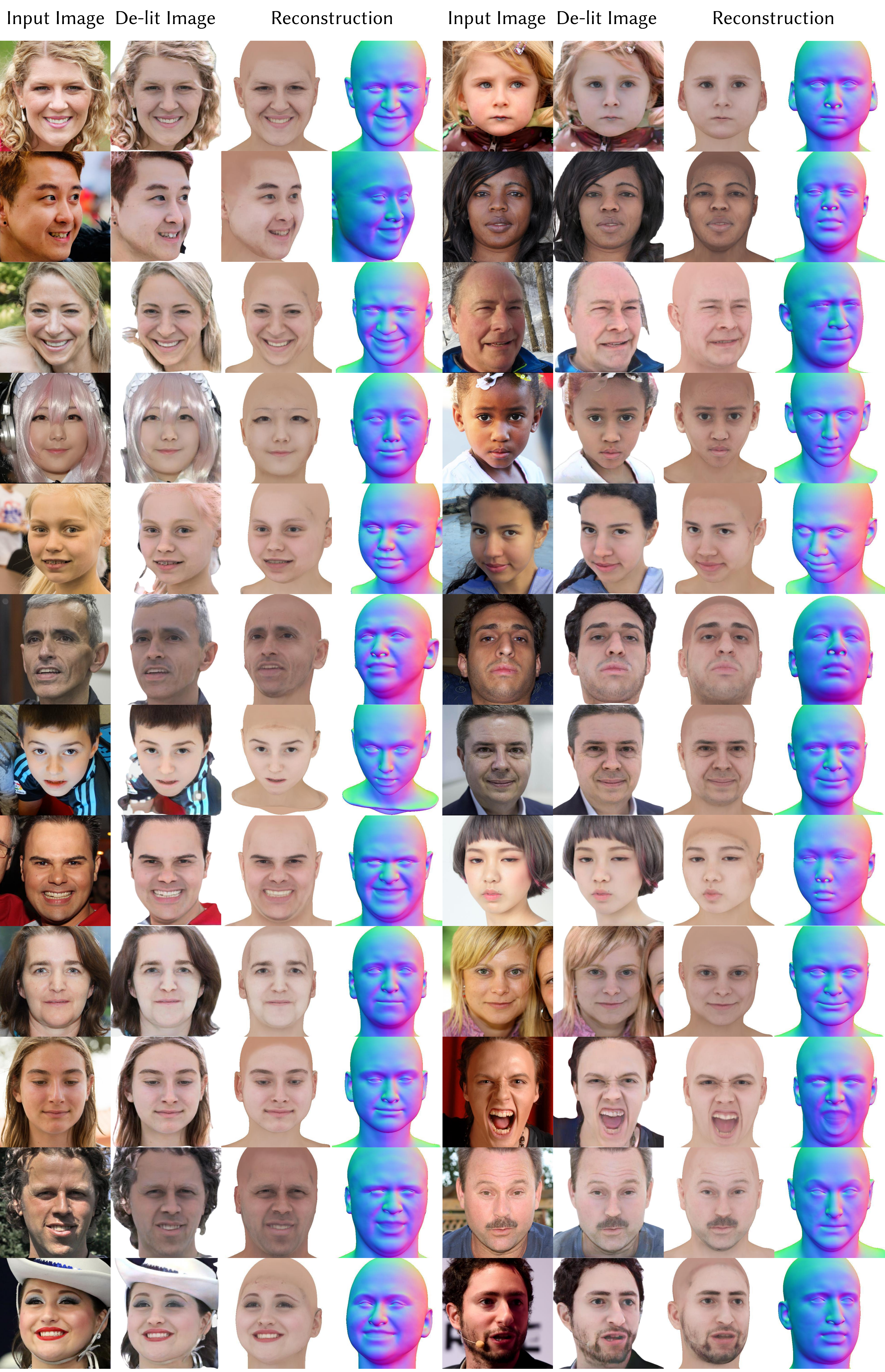}
\end{center}
   \caption{Gallery of single-shot reconstruction results on FFHQ. On the left from top to bottom, images are courtesy of Kerry Goodwin/flickr, Alex "Khaki" Vance/flickr, Katherine Donovan/flickr, Wilson Seed/flickr, SC IPHC/flickr, Commander, U.S. Naval Forces Europe-Africa/U.S. 6th Fleet/flickr, Ordiziako Jakintza Ikastola/flickr, Cena Mineira/flickr, Report Verlag/flickr, Malcolm Slaney/flickr, Gitta Wilén/flickr, and Jill Carlson/flickr. On the right from top to bottom, images are courtesy of Pawel Loj/flickr, Santuario Torreciudad/flickr, Wilbur Ince/flickr, Existence Church/flickr, Eden, Janine and Jim/flickr, Ehud Kenan/flickr, Aécio Neves Presidente/flickr, VcStyle/flickr, Pawel Loj/flickr, Jason Aspinall/flickr, Logan C/flickr, and RISE/flickr.}
   \label{fig:gallery}
\end{figure*}

\end{document}